# Do AI Voices Learn Social Nuances?
# A Case of Politeness and Speech Rate

Eyal Rabin[1], Zohar Elyoseph[2], Rotem Israel-Fishelson[3], Adi Dali[1], and Ravit Nussinson[1]

**Affiliations:** [1]Department of Education and Psychology, The Open University of Israel

[2]Faculty of Education, Department of Counseling and Human Development, Haifa University, Israel [3]College of Education, University of Maryland, United States

## Abstract

Voice-based artificial intelligence is increasingly expected to adhere to human social conventions, but can it learn implicit cues that are not explicitly programmed? This study investigates whether state-of-the-art text-to-speech systems have internalized the human tendency to reduce speech rate to convey politeness - a non-obvious prosodic marker. We prompted 22 synthetic voices from two leading AI platforms (AI Studio and OpenAI) to read a fixed script under both "polite and formal" and "casual and informal" conditions and measured the resulting speech duration. Across both AI platforms, the polite prompt produced slower speech than the casual prompt with very large effect sizes, an effect that was statistically significant for all of AI Studio's voices and for a large majority of OpenAI's voices. These results demonstrate that AI can implicitly learn and replicate psychological nuances of human communication, highlighting its emerging role as a social actor capable of reinforcing human social norms.

# Introduction

Generative artificial intelligence (GenAI) systems now mediate millions of daily conversations, fundamentally reshaping communication by moving interactions with machines from purely functional exchanges to nuanced social encounters. As voice-based agents become deeply integrated into sensitive domains such as healthcare, education, and personal companionship, their ability to navigate complex human social conventions is no longer a mere technical feature, but a fundamental requirement for establishing trust, ensuring user acceptance, and guaranteeing efficacy. This new reality raises a critical question: Do these systems, which learn from vast corpora of human data, implicitly acquire the subtle, non-literal rules that govern social conduct? Can artificial intelligence (AI) learn not just *what* to say, but *how* to say it in a socially appropriate manner? This study addresses these questions by investigating a well-documented yet implicit human behavior - the tendency to reduce speech rate to convey politeness (Nussinson et al., 2025) - as a test case to probe the depth of social learning in state-of-the-art voice AI.

Voice-based AI systems, which operate through spoken language, are becoming more common because speaking and listening are natural forms of interaction, even for users who may not be literate (Carolus et al., 2023). Advances in both language understanding and speech generation have fueled this growth. Large language models (LLMs) enable these systems to understand context and generate comprehensive responses. When combined with speech synthesis, they can produce human-like voice outputs, allowing for more natural conversations between humans and machines.

Text-to-Speech (TTS) systems driven by deep learning can convert written text into speech that sounds exceptionally natural, closely mimicking human speech (Barakat et al., 2024). Current TTS models can produce stylistically modulated speech, for instance, "gender-ambiguous," "formal," or "friendly," based solely on input prompts (Sigurgeirsson & King, 2024; Szekely et al., 2024) or latent conditioning signals such as intonation and rhythm (Barakat et al., 2024). This means synthetic voices are no longer flat or robotic; instead, they can convey tone, emotion, and emphasis much like a human speaker (Abdulrahman & Richards, 2022). Such models can simulate and reproduce complex vocal expressions, for instance, politeness behavior. Such capability is vital as some users attribute social and moral qualities to voice-based agents based on prosodic style (Ribino, 2023). For example, polite-sounding agents are rated as more trustworthy and considerate, even when the semantic content is held constant (Hoegen et al., 2019). However, such modulation often reflects surface-level prosodic mimicry through reproducing statistical correlations between text style and vocal delivery from training data, without genuine understanding of the social context or pragmatic reasoning (e.g., Liu et al., 2021; Wang et al., 2018). This raises important theoretical and empirical questions about the nature of the social knowledge that these systems are encoding.

Various empirical studies on humans provide evidence for a politeness–speech rate association, such that slower (/faster) speech is associated with more polite (/more casual) speech, with a broad consensus of findings across different languages. For instance, Japanese speakers were found to speak more slowly when using honorific (polite) forms as compared to when using casual forms (Ofuka et al., 2000), and Catalan speakers similarly used a slower rate when making requests of high imposition or addressing unfamiliar listeners (Staszkiewicz, 2024). A recent series of studies found that participants experience a slower (/faster) version of a message in a foreign language as more compatible with a polite (/casual) message; that participants intended to speak more slowly (/faster) when they wished to speak more politely (/in a casual manner); and that participants actually spoke more slowly when they intended to be more polite (Nussinson et al., 2025).

This paper investigates whether advanced voice-based AI systems also exhibit this latent social cue, specifically that they speak more slowly when speaking politely compared to when speaking casually. We compare two cutting-edge TTS systems (one developed by OpenAI and the other by Google) in their ability to reproduce the well-attested human pattern linking politeness with a reduced speech rate. By examining how each system modulates speech tempo under polite vs. casual conditions, we aim to shed light on how well machines have implicitly learned human pragmatic norms. Before describing our empirical approach, we review the relevant literature from three intersecting domains: (1) GenAI and voice-based communication, (2) human perceptions of synthetic voices in Human-Machine Interactions, and (3) core psychological theories of politeness and speech rate. This integrated review will ground our hypotheses and highlight the theoretical significance of testing whether AI voices can simulate latent human social norms in speech.

**Generative AI and Voice-based Communication**

The emergence of GenAI and LLMs has led to significant breakthroughs in voice-based communication, blurring the lines between machine output and human expression (Sigurgeirsson & King, 2024). Synthetic speech in TTS systems, which was once monotonous and robotic, has evolved and improved considerably over the past years (Mehrish et al., 2023).

Early advancements utilizing deep learning and neural architectures, such as Tacotron 2 (Hussen Abdelaziz et al., 2021) and WaveNet (Shen et al., 2018), allowed for end-to-end mapping from text to audio, producing natural-sounding speech (Ning et al., 2019). However, these solutions were not optimal in terms of support for prosody and expression. Newer innovations based on LLMs have further enhanced the ability to generate fluent, personalized, and contextually expressive speech (Lakomkin et al., 2024).

One notable advance that GenAI models have introduced is the ability to control prosody and add expressive variations in synthetic speech. Prosody refers to qualities of speech such as rhythm, pitch, stress, and intonation patterns, which are crucial for conveying nuanced meanings (Cole, 2015). These nuances are very important because they convey emotional and pragmatic information that cannot be expressed with a monotone voice. Recent models and TTS systems offer explicit mechanisms for prosodic conditioning and control, enabling nuanced modulation of tone, rhythm, and emotion. NaturalSpeech 3 (Ju et al., 2024) is an example of such a system that breaks down speech into distinct layers, including voice, identity, and prosody, while allowing for changes in tone or rhythm without compromising meaning. Other TTS systems enable fine-grained prosodic control through different methods and features. A recent systematic review offers a comprehensive analysis of how prosody has been modeled and evaluated in contemporary TTS research (Galdino et al., 2025). Their review, which synthesizes 100 peer-reviewed studies, sheds light on dominant prosodic parameters and highlights the need for evaluation strategies to properly assess prosodic fidelity. A similar effort was undertaken by Barakat et al. (2024), who explored deep-learning expressive approaches, including both supervised and unsupervised methods, while addressing challenges related to prosodic control. These technological advances lay the foundation for voice-based AI systems that are not only intelligible but also socially expressive, enabling applications that extend far beyond simply delivering content. Such applications can be seen across various fields. In education, for example, such technologies are integrated into bots that assist students in learning foreign languages (Tai & Chen, 2024). In customer service, AI-powered voice agents are utilized in call centers to address various inquiries, reducing customer complaints (Wang et al., 2023). The proliferation of these technologies and their impact on our lives raises important issues about their social, ethical, and psychological implications for human-machine interaction and communication.

**Perceptions of Synthetic Speech in Human-Machine Interactions**

The human perception of AI-based synthetic speech systems has been studied in various settings and with different populations (e.g., Herrmann, 2023; Ross et al., 2024). Acoustic and contextual factors, including emotional tone, interpersonal intent, and social cues, influence these perceptions (Fan & Liu, 2025).

On one hand, the growing human-like nature of AI-based synthetic speech fosters anthropomorphic perceptions. Users often attribute personality traits and emotional states to machines based on prosody and vocal cues, even in the absence of physical embodiment (Ehret et al., 2021; Yang et al., 2024). Various prosodic qualities such as tone, pitch, and rhythm were found to contribute to impressions of warmth, credibility, competence, or friendliness (Rallabandi et al., 2021; Seaborn et al., 2021). These qualities influence the acceptance of AI technologies, willingness to engage, and trust (Choung et al., 2023; Fan & Liu, 2025).

On the other hand, mismatches between the prosody of synthetic speech and user expectations may elevate cognitive load and reduce satisfaction (Delogu et al., 1998). The concept of the Uncanny Valley, coined by Mori et al. (2012), describes the discomfort experienced by viewers when synthetic entities do not fully replicate their human counterparts. Today, it is accepted that the phenomenon applies not only to humanoid robots but also to voices. When synthetic speech is nearly human in prosody, but contains subtle irregularities, it can evoke discomfort or distrust (Do et al., 2022). From a psychological perspective, the discomfort could arise from expectancy violations and cognitive dissonance. Listeners expect an alignment between the content discussed and the voice's qualities. A mismatch between the two would lead to an unsettling feeling. Ongoing research suggests that by carefully manipulating specific speech features, such as pitch, intonation, rhythm, and clarity, in synthesized voices, it is possible to make them more acceptable and even preferred in certain contexts. Improving these features could enhance human-robot interaction, making it more natural and comfortable for users, and potentially increasing the effectiveness of assistive robots, virtual assistants, and other automated systems (Kühne et al., 2020).

**Core Psychological Theories of Politeness and Speech Rate**

Speech rate is a prominent parameter in both human and synthetic communication, influencing cognitive processing and emotional evaluations. Long-lasting research in psycholinguistics suggests a link between speech rate and persuasiveness, empathy, and fluency. While slow speech may be perceived as empathetic and considerate, fast speech and raised pitch are associated with persuasiveness but also with irritability (Apple et al., 1979).

Cognitive load theory (Sweller, 1988) offers a useful framework for understanding how speech rate affects working memory and comprehension. Cognitive load can be caused by the complexity of the content delivery and by a mismatch between the speech rate and the processing capacity. A faster rate makes it harder to decode syntactic structure and phonological encoding, while a slower rate increases the interval between information units and may thus require more cognitive resources to maintain context. Both extremes can reduce processing fluency and increase listening effort (Colby & McMurray, 2021). A high cognitive load caused by inappropriate speech rates can impair understanding and memory retention, particularly with complex or unfamiliar topics. However, adaptive strategies, such as pausing at clause boundaries and using prosodic cues to mark important content, can mitigate these effects (Beier et al., 2025).

Speech rate not only influences cognitive processing but also serves as a subtle yet powerful cue in the management of social interactions. The most influential framework for understanding this process is Politeness Theory, developed by Brown and Levinson (1987). The theory posits that speakers are motivated to protect their own and their interlocutor's "face" - the public self-image that every person wants to claim. Many speech acts, such as making a request, constitute a Face-Threatening Act (FTA) because they impose on the hearer's autonomy. To mitigate these threats, speakers employ politeness strategies. A slower speech rate can serve as a key component of such strategies, particularly "negative politeness." By speaking more slowly, a speaker can signal deference, reduce the perceived imposition of the request, and convey that they are not rushing or pressuring the listener, thereby preserving social harmony (Yusupova, 2025). Furthermore, recent findings suggest that slow speed is associated with psychological distance and that, more specifically, slow-pace speech is associated with greater social distance between the speaker and the interlocutor (Nussinson et al., 2024). As politeness is known to both reflect and create social distance (Stephan et al., 2010), slow-paced speech may be associated with politeness exactly because politeness is a manifestation of social distance (Nussinson et al., 2025). This theoretical lens provides a direct rationale for the hypothesis that polite speech is systematically associated with a reduced tempo.

**Politeness in AI**

As AI agents become an integral part of social life, their ability to adhere to human norms of politeness is critical for fostering user trust, acceptance, and effective collaboration (Ribino, 2023). Early research in Human-Computer Interaction, particularly the "Computers as Social Actors" (CASA) paradigm, established the understanding that users naturally apply social rules to machines and respond to cues of politeness or impoliteness (Reeves & Nass, 1996). Consequently, a significant portion of the work on politeness in AI has focused on implementing explicit politeness strategies, primarily through lexical and syntactic choices. This includes programming agents to use words like "please" and "thank you," often driven by concerns that command-based interactions with digital assistants could negatively affect social behavior, especially in children (Burton & Gaskin, 2019). However, focusing solely on lexical markers overlooks the primary channel through which social meaning is conveyed: prosody. Authentic social competence requires more than adherence to explicit rules; it involves mastering the subtle, non-verbal cues that often accompany and even override verbal content. Prosody—the rhythm, pitch, and rate of speech—is a central channel for this implicit social signaling (Luo, 2025).

These failures highlight a critical question: if AI systems struggle even to adapt explicit politeness strategies to different cultural contexts, can they implicitly learn and internalize the subtle, undefined prosodic cues from the data patterns on which they were trained? This study addresses this gap by investigating a specific, implicit prosodic cue—the slowing of speech

rate—to examine whether advanced generative AI has acquired a nuanced understanding of human social norms from statistical data, without explicit instruction.

## Methods

### *Materials and Procedure*

The study employed a fully crossed design in which two text-to-speech (TTS) systems (AI Studio by Google vs. OpenAI) were each evaluated under two speaking-style conditions ("Casual and Informal" vs. "Polite and Formal"). For each system, we selected 11 distinct synthetic voices; each voice was prompted to produce the target script 10 times in each style condition, yielding a total of 2 systems × 2 styles × 11 voices × 10 utterances = 440 recordings.

### *Target Script.*

A single 105-word passage was used across all voices and conditions:

*"Hi, my name is [Gemini / OpenAI]. In this study we need to get acquainted with each other. First, I will tell you a little bit about myself. After I finish, I would be glad if you could answer some questions. I hope you do not mind filling out a questionnaire regarding your preferences in various areas. I already filled out a similar questionnaire, so we can look at them and get a clue about each other before we start talking. When we begin to talk, I will ask you some questions first, and then you can ask me. I would appreciate it if you would tell me if you don't feel comfortable with the questions I will ask."*

*Instruction Prompts*

- *Casual and Informal Prompt.*

  *"Please record yourself addressing the student by reading the text below, casually and informally (without changing the text). Try to speak in a casual and informal manner."*

- *Polite and Formal Prompt.*

  *"Please record yourself addressing the student by reading the text below, politely and formally (without changing the text). Try to speak in a polite and formal manner."*

*Audio Generation*

For each system, we iterated through its 11 available voices. In each style condition, the corresponding instruction prompt and the target script were submitted to the TTS interface. Each voice-condition pair was sampled ten times (with identical prompt text but distinct generation seeds), producing ten unique renditions per pairing.

*Randomization and Export*

The order of voice and style presentation was fully randomized separately for each system to control for potential order effects. All audio outputs were generated at a 24 kHz sampling rate and exported as .WAV files, then stored with filenames indicating system, voice ID, style condition, and sample number (e.g., "AIStudio_Voice03_Casual_07.wav").

*Quality Check*

Following generation, each recording was inspected for completeness (i.e., absence of synthesis errors or truncation). Any flawed samples ($< 1\%$ of total) were regenerated immediately. This procedure ensured balanced coverage of voices and speaking styles across both TTS systems, facilitating a comprehensive comparison of their vocal performance under casual versus formal speaking-style conditions.

### *Data and Materials Availability*

All audio files generated for this study, as well as the data file containing the measured speech durations for each recording, are publicly available on the Open Science Framework (OSF) at the following link: https://osf.io/nyqae/?view_only=018e61edd2574ee89cc460f0988c6fb1

## Results

All statistical analyses were conducted using SPSS v.29. Prior to hypothesis testing, reaction-time data were screened for outliers defined as values exceeding ±3 *SD* from the group mean (Tabachnick & Fidell, 2013). Homogeneity of variance between the polite and casual conditions was assessed via Levene's test for each voice (Levene, 1960).

To evaluate the effect of phrasing (polite vs. casual) on response time for each AI Studio and OpenAI voice, we performed independent-samples *t*-tests separately for each of the 22 voices. Given the large number of comparisons and the attendant risk of inflated Type I error, raw *p*-values were adjusted using the Holm–Bonferroni procedure (Holm, 1979). Holm's sequentially rejective method orders *p*-values from smallest to largest and compares each to a threshold of α/(m – k + 1), where *m* is the total number of tests and *k* is the rank of the *p*-value, thus controlling the family-wise error rate with greater power than the simple Bonferroni correction (Abdi, 2007).

Effect sizes for each comparison were calculated as Cohen's *d*, using the pooled standard deviation (Cohen, 1988). Ninety-five percent confidence intervals and Cohen's *d* are presented to aid interpretation of effect magnitude and precision (Cumming, 2012).

For both AI Studio and OpenAI, speech durations for polite versus casual phrasing across the 11 voices were compared using independent-samples *t*-tests, with *p*-values adjusted via the Holm–Bonferroni procedure to control family-wise error (Holm, 1979). Tables 1 and 2 present the adjusted *p*-values, *t*-statistics, degrees of freedom, and Cohen's *d* effect sizes for each voice. Statistical significance was evaluated at α = .05 (one-tailed) after correction. Table 1 presents the results of AI Studio, and Table 2 presents the results for OpenAI. The findings indicate a consistent trend across both platforms. For AI Studio, all voices produced significantly slower speech in response to polite phrasing. For OpenAI, this effect was significant for 8 out of 11 voices, while the remaining voices also exhibited a slower, albeit non-significant, speech rate in response to the polite prompt.

**Table 1**

Independent-Samples *t*-Tests Comparing Polite and Casual Conditions for AI Studio Voices

| Voice | Polite M(SD) | Casual M(SD) | t | p | Cohen's d | 95% CI [LL, UL] |
|---|---|---|---|---|---|---|
| Aoede | 43.14 (3.42) | 37.07 (1.42) | 5.18 | < .001 | 2.32 | [3.60, 8.52] |
| Autonoe | 46.42 (3.07) | 37.17 (2.19) | 7.76 | < .001 | 3.47 | [6.74, 11.75] |
| Callirrhoe | 47.95 (5.76) | 37.87 (4.06) | 4.52 | < .001 | 2.02 | [5.39, 14.75] |
| Charon | 48.41 (4.76) | 32.70 (2.17) | 9.48 | < .001 | 4.24 | [12.23, 19.30] |
| Fenrir | 44.46 (5.02) | 34.60 (2.18) | 5.69 | < .001 | 2.55 | [6.22, 13.49] |
| Kore | 44.18 (3.78) | 36.06 (2.05) | 5.97 | < .001 | 2.67 | [5.26, 10.98] |
| Leda | 44.43 (3.07) | 36.90 (1.87) | 6.62 | < .001 | 2.96 | [5.13, 9.91] |
| Orus | 42.63 (3.72) | 37.23 (4.86) | 2.78 | .01 | 1.24 | [1.32, 9.46] |
| Puck | 46.22 (5.39) | 35.68 (3.05) | 5.39 | < .001 | 2.41 | [6.44, 14.65] |
| Sulafat | 47.48 (5.88) | 36.29 (2.93) | 5.39 | < .001 | 2.41 | [6.83, 15.56] |
| Zephyr | 46.92 (3.97) | 38.19 (3.09) | 5.48 | < .001 | 2.45 | [5.38, 12.07] |

*Note*. N = 10 for each voice in each condition. Speech duration is measured in seconds. CI = Confidence Interval; LL = Lower Limit; UL = Upper Limit. All *t*-tests have 18 degrees of freedom. *p*-values are adjusted using the Holm-Bonferroni procedure.

**OpenAI fm**

Speech durations for polite versus casual phrasing across the eleven OpenAI fm voices were compared using independent-samples t-tests, with p-values adjusted via the Holm–Bonferroni procedure to control family-wise error (Holm, 1979). Levene's tests confirmed homogeneity of variances in all cases (all p > .05). Table 2 presents the Holm–Bonferroni–adjusted p-values, t-statistics, degrees of freedom, and Cohen's d effect sizes for each voice. Statistical significance was evaluated at α = .05 (one-tailed) after correction.

**Table 2**

Independent-Samples t-Tests Comparing Polite and Casual Conditions for OpenAI fm Voices

| Voice | Polite M(SD) | Casual M(SD) | t | p | Cohen's d | 95% CI [LL, UL] |
|---|---|---|---|---|---|---|
| Fable | 35.44 (1.26) | 32.74 (1.03) | 5.22 | < .001 | 2.34 | [1.61, 3.78] |
| Verse | 36.08 (3.02) | 32.49 (1.33) | 3.44 | .006 | 1.54 | [1.39, 5.78] |
| Alloy | 36.58 (4.51) | 33.77 (4.32) | 1.42 | .172 | 0.64 | [−1.34, 6.96] |
| Ash | 35.55 (2.90) | 33.02 (1.17) | 5.35 | < .001 | 2.39 | [1.53, 3.51] |
| Ballad | 35.75 (0.98) | 33.95 (1.49) | 3.19 | .015 | 1.43 | [0.62, 2.99] |
| Coral | 34.82 (2.44) | 32.68 (2.04) | 2.12 | .072 | 0.95 | [0.02, 4.26] |
| Echo | 35.42 (1.37) | 32.90 (0.77) | 5.05 | < .001 | 2.26 | [1.47, 3.56] |
| Nova | 34.75 (0.54) | 31.21 (1.06) | 9.40 | < .001 | 4.21 | [2.75, 4.33] |
| Onyx | 34.10 (2.10) | 33.21 (1.23) | 1.15 | .133 | 0.56 | [−0.74, 2.50] |
| Sage | 37.30 (1.36) | 34.36 (1.19) | 5.13 | < .001 | 2.20 | [1.74, 4.14] |
| Shimmer | 34.26 (2.37) | 31.94 (1.32) | 2.70 | .028 | 1.21 | [0.52, 4.13] |

*Note.* $N = 10$ for each voice in each condition. Speech duration is measured in seconds. *CI* = Confidence Interval; LL = Lower Limit; UL = Upper Limit. All *t*-tests have 18 degrees of freedom. *p*-values are adjusted using the Holm-Bonferroni procedure.

**Discussion**

Human interaction is saturated with subtle social cues that often transcend explicit verbal content, enabling effective communication management. This study investigated whether modern voice-based AI systems have internalized one such cue: the empirically established human tendency to slow down speech rate to convey politeness (Ofuka et al., 2000; Nussinson et al., 2025). Crucially, speech rate is a non-obvious prosodic feature, one that is unlikely to be explicitly programmed into voice models. The core question, therefore, was whether AI can learn such implicit, psychological nuances, in this case, the translation of a social concept like politeness into a specific acoustic modification. The findings were replicated across two different platforms. The effect was statistically significant for all of AI Studio's voices and for the majority of OpenAI's voices, with the remaining voices showing a similar, non-significant, pattern. In all cases, the synthetic voices tested produced slower speech when prompted to speak in a "polite and formal" manner compared to a "casual and informal" one. This result, demonstrated through consistently large effects, provides strong evidence that large language models are capable of learning and replicating these subtle prosodic patterns, even without explicit instruction.

The most plausible mechanism underlying this finding is not a genuine "understanding" of politeness, but rather a process of "stochastic parrots" (Bender et al., 2021). Trained on vast datasets of human speech, the systems have likely identified the correlation between lexical markers of politeness (e.g., the use of words like "please" or "thank you" in texts) and accompanying acoustic features, such as a slower speech rate. In doing so, they have learned to associate the prompt "be polite" with the correct prosodic feature. This finding extends the Computers as Social Actors (CASA) paradigm (Reeves & Nass, 1996), showing that machines are not only perceived by us as social actors but are also becoming increasingly capable of acting in a manner consistent with social norms, even if their actions are based on pattern recognition rather than social intent.

It is interesting to compare the magnitude of the effect observed in the AI systems to those found in parallel studies on humans. The effect sizes in the current study (Cohen's d) were very large, in many cases exceeding those observed in an identically parallel study in human behavior (e.g., Nussinson et al., 2025, Study 3). This may suggest that AI, having learned the statistical rule, applies it more consistently, and perhaps even in a more exaggerated manner, than humans, whose behavior is influenced by a wider range of contextual and personal variables. Furthermore, the differences in effect sizes between the Google system (where the effects were particularly large) and the OpenAI system indicate that there may be significant variations between different models, possibly resulting from differences in training data or model architecture.

The implications of these findings are potentially far-reaching. Voice-based AI systems mediate millions of interactions daily. When these systems replicate human social cues, they not only enhance the naturalness of the interaction but also become active partners in shaping and reinforcing social norms. Every interaction where a digital assistant "chooses" to slow its speech to sound polite reinforces the user's cultural association between slowness and politeness. This phenomenon transforms AI from a mere technological tool into an agent with a socializing influence.

This study has several limitations that provide a basis for future research. First, we used a single script and limited style prompts. Future studies could examine a wider variety of social contexts, scripts, and languages. Second, we focused on a single prosodic measure (speech rate), whereas politeness is conveyed through a combination of cues such as pitch, intonation, and volume (Brown, & Prieto, 2017). Another fascinating research direction would be to examine the gap between the system's "procedural" (behavioral) knowledge, as demonstrated here, and its "declarative" (stated) knowledge. In other words, it would be worthwhile to examine how AI explicitly responds to a question about the connection between speech rate and politeness. Beyond these specific avenues, the methodology presented here offers a robust framework for comparing the vocal capabilities of LLMs with human baselines, providing a valuable tool for future research in human-AI interaction.

In conclusion, this research demonstrates that generative AI is beginning to acquire abilities that go beyond mere language processing, adopting subtle features of human social behavior. Although this ability is likely the product of sophisticated mimicry rather than deep social understanding, it has significant implications for the future of human-machine interaction and for how we understand social learning processes. Machines are learning not only what we say, but also the unwritten rules of how we say it.

**Acknowledgments**

None

**Funding**

Anonymized for review

**Conflict of Interest Statement**

The authors declare that there are no conflicts of interest with respect to the research, authorship, and/or publication of this article.

**Declaration of generative AI and AI-assisted technologies in the manuscript preparation process.**

During the preparation of this work the authors utilized Google's Gemini and OpenAI's ChatGPT 4 to enhance the clarity and phrasing of the manuscript and to aid in structuring the content. Following the use of these services, the authors conducted a thorough review and editing process and assume full responsibility for the final content of the published article.